\documentclass{article}
\usepackage{amsmath} \usepackage{amssymb,bm}
\usepackage{pdfsync}
\usepackage{graphicx}

\newcommand{\mydef}{\triangleq}
\newcommand{\E}{\mathbf{E}}
\newcommand{\ma}[1]{ \begin{bmatrix} #1 \end{bmatrix} }
\newtheorem{thm}{Theorem}
\newtheorem{cor}{Corollary}
\newenvironment{pf}{
    \par
\noindent
    {\bf Proof:}
    \begingroup 
  }
  {
    \endgroup
        \ \hfill $\Box$ \par\bigskip
  }

  \title{Analysis of Interpolating Regression Models and the Double
  Descent Phenomenon\footnote{Accepted for presentation at IFAC
    World Congress, Yokohama, Japan, July 2023}} 

\author{Tomas McKelvey\\Chalmers University of Technology, 
   Gothenburg, Sweden \\ (tomas.mckelvey@chalmers.se) } 

\begin{document}

\maketitle
\begin{abstract}                
  A regression model with more parameters than data points in the
  training data is overparametrized and has the capability to
  interpolate the training data. Based on the classical bias-variance
  tradeoff expressions, it is commonly assumed that models which
  interpolate noisy training data are poor to generalize. In some
  cases, this is not true. The best models obtained are
  overparametrized and the testing error exhibits the double descent
  behavior as the model order increases.  In this contribution, we
  provide some analysis to explain the double descent
  phenomenon, first reported in the machine learning literature. We
  focus on interpolating models derived from the minimum norm solution
  to the classical least-squares problem and also briefly discuss
  model fitting using ridge regression. We derive a result based on
  the behavior of the smallest singular value of the regression matrix
  that explains the peak location and the double descent shape of the
  testing error as a function of model order.
\end{abstract}

\section{Introduction}
\label{sec:intro}
Linearly parametrized regression models that interpolate the training data have recently
attracted significant attention
\cite{10.1214/21-AOS2133}, mainly due to the close connections
to many state-of-the-art machine learning  models \cite{https://doi.org/10.48550/arxiv.1611.03530,belkin2019reconciling}. Interpolation of
the training data is
obtained when the number of estimated/trained parameters in the model is equal to
or larger than the number of training data used to estimate the model. Such models are hence
overparametrized and there exists an infinite number of solutions that
interpolate the data. It has been noticed in recent publications
\cite{10.1214/21-AOS2133,ribeiro2021beyond} that the quality of an
estimated model with a linear parametrization often follows a so-called
double descent curve. This means that the test data error first
decreases with an increasing model order and then increases again to a maximum
when the number of samples of the data is equal to the number of
parameters and then gradually decreases again with an increasing model
order. The double descent phenomenon has previously been noticed in
the deep-learning area \cite{belkin2019reconciling,loog2020brief}. 

In this paper, we will show that this behavior can be explained
with a minimum of theory and complement the available literature
on the subject. Furthermore, we point out that the extent of the
behavior is closely tied to how the model class is constructed, i.e.,
the ordering of the basis functions employed in the regression model.  

The paper is structured as follows. In Section~\ref{sec:probform}, the
regression problem is formulated and necessary notation and assumptions
are explained. In Section~\ref{sec:analysis}, we derive explicit expressions
for the bias and variance contributions for predictions obtained using
the estimated model and show that the smallest singular value of the
regression matrix plays a key role. We further provide a theorem that
predicts the behavior of the smallest singular value as the model
order increases. A few numerical examples are given in
Section~\ref{sec:numer-illustr} that illustrate the connection
between the model quality and the smallest singular value. In Section~\ref{sec:conclusions}, the paper
is concluded with a summary of the findings.
\section{Problem formulation}
\label{sec:probform}
We consider a general regression problem where we seek a function
that maps a value in the domain $\mathcal{X}$ to the co-domain
$\mathcal{Y}$, i.e. $f: \mathcal{X} \rightarrow \mathcal{Y}$ based on
a given set of samples of training data
\begin{equation}
  \label{eq:Dset}
  \mathcal{D} = \{ (y(t), x(t)) \}_{t=1}^{M}
\end{equation}
where $x(t)\in\mathcal{X}$ and $y(t)\in\mathcal{Y}$
and we consider a linearly parametrized function class
\begin{equation}
  \label{eq:flinpar}
  f(x;\bm \theta) = \sum_{i=1}^n \theta_i \phi_i(x) = \bm{\phi}(x) \bm{\theta}
\end{equation}
where $\phi_i: \mathcal{X} \rightarrow \mathcal{Y}$ are given distinct
basis functions and
\begin{equation}
  \label{eq:bmphi}
  \begin{aligned}
    \bm\phi(x) & =
    \begin{bmatrix}
      \phi_1(x) &  \phi_2(x) & \cdots & \phi_n(x) 
    \end{bmatrix} \\
    \bm\theta & =
    \begin{bmatrix}
      \theta_1 &  \theta_2 & \cdots & \theta_n 
    \end{bmatrix}^T \in\mathcal{T}. 
  \end{aligned}
\end{equation}
The model order is equal to the size of the parameter vector $\bm
\theta$ and is denoted by the integer $n$. In the analysis that
follows we assume the sets $\mathcal{X}$ and $\mathcal{Y}$ can be real or
complex spaces with finite dimensions $m$ and $p$ respectively and the parameter set
$\mathcal{T}$ can be real or complex valued with finite dimension $n$.

Based on the training data set the model parameters are determined by
minimizing  the sum of squared errors 
\begin{equation}
  \label{eq:lsproblem}
  \hat{\bm{\theta}} = \arg \min_{\bm{\theta}} \sum_{t=1}^{M}\|y(t)- \sum_{i=1}^n \theta_i \phi_i(x(t)) \|^2
\end{equation}
The minimization problem above can be written as the least-squares
(LS) problem
\begin{equation}
  \label{eq:LSmatrix}
   \hat{ \bm \theta} = \arg \min_{\bm\theta} \| \bm{y} - \bm{\Phi} \bm\theta \|^2 
 \end{equation}
 with the \emph{regression matrix}
 \begin{equation}
   \label{eq:Phi}
   \bm{\Phi} =
   \begin{bmatrix}
     \bm\phi(x(1)) \\
     \bm\phi(x(2)) \\
     \vdots \\
          \bm\phi(x(M)) \\
        \end{bmatrix}
        = 
   \begin{bmatrix}
     \phi_1(x(1))  & \phi_2(x(1)) & \cdots & \phi_n(x(1)) \\   
     \phi_1(x(2)) & \phi_2(x(2)) & \cdots & \phi_n(x(2)) \\
     \vdots & \vdots & \vdots &\vdots \\
     \phi_1(x(M)) & \phi_2(x(M)) & \cdots & \phi_n(x(M)) \\
   \end{bmatrix}
 \end{equation}
 which is a matrix with $N \mydef  p M$ rows and $n$ columns and the
 vector 
 \begin{equation}
   \label{eq:ybf}
   \bm{y} =
   \begin{bmatrix}
     y(1) \\
     y(2) \\
     \vdots \\
     y(M)
   \end{bmatrix}. 
 \end{equation} has $N$ elements.
We note that if $n\leq N$ we have an under-parametrized problem and
the solution to \eqref{eq:LSmatrix} is unique if $\mathbf{\Phi}$ has
full rank. If $n>N$ the problem is overparametrized and there exist
infinite many solutions to \eqref{eq:LSmatrix}. 
 The singular value decomposition (SVD) \cite{Golub&VanLoan:89} of the regression matrix
$\mathbf{\Phi}$  in \eqref{eq:bmphi} plays a key
role in the analysis in this paper and we denote it as
 \begin{equation}
   \label{eq:PhiSvd}
   \mathbf{\Phi} = \sum_{k=1}^r \sigma_k\bm{u}_k \bm{v}_k^H
 \end{equation}
 where $r=\min(n,N)$, the
 ordered singular
 values are $\sigma_1\geq \sigma_2 \geq \ldots \geq \sigma_r  \geq 0
 $, and $\bm{u}_k$ and $\bm{v}_k$ are the left and right singular vectors
 respectively and $(\cdot)^H$ denote the Hermitian transpose. We assume 
 $\bm{\Phi}$ has full rank and hence $\sigma_r>0$.

For analysis purposes we assume there exists a function $f_0:
\mathcal{X} \rightarrow \mathcal{Y}$ such that the data generated can be described as
 \begin{equation}
   \label{eq:datagen}
   y(t) = f_0(x(t)) + z(t) 
 \end{equation}
where $z(t)$ is an i.i.d.\ zero mean white noise process with
variance $$\E \, z(t) z(t)^H \mydef \bm{R}_z.$$

\section{Analysis}
\label{sec:analysis}
In this section, we discuss and provide some analytical results on the
bias and variance of the estimated model that
is given by the minimum norm solution and the ridge regression
solution.  We show that the largest variance contribution
is proportional to $1/\sigma_r^2$, the inverse of the square of the smallest
singular value of the regression matrix $\bm\Phi$. Finally, we show
that when $n<N$ and the model order is increased to $n+1$ the smallest singular
value will decrease or stay unchanged. This implies that the variance
increases or stays constant with the increase of the model
order. Furthermore, when $n\geq N$ and the model order is
increased to $n+1$ we show that the smallest singular value is increased or
stays the same. This implies that the variance
decreases or stays constant with an increase in the model order.
\subsection{The minimum norm solution}
\label{sec:minim-norm-solut}
The unique \emph{minimum 
 norm solution} to \eqref{eq:LSmatrix} is obtained with the
Moore-Penrose pseudo-inverse and can be expressed using the SVD
as (see, e.g.\  \cite{Golub&VanLoan:89, laub2005matrix}) 
 \begin{equation}
   \label{eq:min_norm_sol}
   \bm{\hat \theta} = \bm{\Phi}^+\bm{y} = \sum_{k=1}^r \frac{1}{\sigma_k}\bm{v}_k \bm{u}_k^H \bm{y}.
 \end{equation}
By introducing the notation 
\begin{equation}
  \bm{x} = \ma{x(1)\\x(2) \\ \vdots \\ x(M)}, \;
  \label{eq:1}
  \bm{f}_0(\bm{x}) =
  \begin{pmatrix}
    f_0(x(1)) \\ f_0(x(2)) \\ \vdots \\ f_0(x(M)) 
  \end{pmatrix}, \;
     \bm{z} =
   \begin{pmatrix}
     z(1) \\
     z(2) \\
     \vdots \\
     z(M)
   \end{pmatrix}
\end{equation}
the estimate in \eqref{eq:min_norm_sol} can be decomposed into
\begin{equation}
  \label{eq:2}
  \bm{\hat \theta} = \sum_{k=1}^r \frac{1}{\sigma_k}\bm{v}_k
  \bm{u}_k^H (\bm{f}_0(\bm{x}) + \bm{z}) =
  \bm \theta_* + \tilde {\bm \theta} 
\end{equation}
where $\bm \theta_* = \bm{\Phi}^+\bm{f}_0(\bm{x}) = \sum_{k=1}^r \frac{1}{\sigma_k}\bm{v}_k
  \bm{u}_k^H \bm{f}_0(\bm{x})$ is the noise free minimum
  norm solution and $\tilde {\bm \theta} = \sum_{k=1}^r \frac{1}{\sigma_k}\bm{v}_k
  \bm{u}_k^H \bm{z} $ is the contribution due to the
  noise. The zero mean assumption
  on $z(t)$ yields 
  \begin{equation}
    \label{eq:3}
    \E \,\bm{\hat \theta}  = \bm \theta_*
  \end{equation}
 The error of the output $f(x' ; \hat{\bm\theta})$ of the estimated
 model given a new test data sample $x'$ is 
 \begin{equation}
   \label{eq:prederror}
   \begin{aligned}
     e(x') & = \bm{\phi}(x') \hat{\bm \theta} - f_0(x') \\ & = \bm{\phi}(x')
     \bm \theta_* - f_0(x') + \bm{\phi}(x') \sum_{k=1}^r
     \frac{1}{\sigma_k} \bm v_k \bm u_k^H \bm{z}
   \end{aligned}
 \end{equation}
 From above it is clear that the error $e(x')$ is composed
 of a bias part $\bm{\phi}(x')  \bm
   \theta_* - f_0(x')$ and a zero mean stochastic part  $\bm{\phi}(x') \sum_{k=1}^r \frac{1}{\sigma_k} \bm{v}_k \bm{u}_k^H \bm{z}$  that
 contributes with variance  
 \begin{equation}
   \label{eq:perr_cov}
   \begin{aligned}
    & R_e(x')  \mydef \operatorname{cov}(e(x') ) = \\
     &  \bm{\phi}(x') \left[\sum_{k=1}^r
       \frac{1}{\sigma_k}\bm{v}_k \bm{u}_k^H \right] (I_N \otimes R_z)
          \left[\sum_{k=1}^r \frac{1}{\sigma_k}\bm{v}_k \bm{u}_k^H\right ]^H
     \bm{\phi}^H(x')
   \end{aligned}
 \end{equation}
 If the measurement noise is uncorrelated between the channels and have
 equal variance $r_z$, then $R_z= r_z \bm{I}$. In this case, the
 covariance expression \eqref{eq:perr_cov} is simplified to
 \begin{equation}\label{eq:simp_pecov}
   \begin{aligned}
     R_e(x') & = r_z \sum_{k=1}^{r} \frac{1}{\sigma^2_k} (\bm{\phi}(x')
     \bm{v}_k) (\bm{\phi}(x') \bm{v}_k)^H \\
     & = r_z \bm{\phi}(x') \left( \sum_{k=1}^{r} \frac{1}{\sigma^2_k} 
     \bm{v}_k\bm{v}_k^H
     \right)  \bm{\phi}(x')^H
   \end{aligned}
 \end{equation}
 since $\bm{u}_k^H \bm{u}_l = 0$ for $k\not = l$.
From \eqref{eq:prederror} (and \eqref{eq:simp_pecov}) it is clear that if the smallest singular value of $\bm \Phi$ is close
to zero the error (and variance) can become arbitrarily large unless
$\bm{\phi}(x')$  is perpendicular to the right singular vector
corresponding to the smallest singular value. Further we note that for
any singular value distribution we have the inequality
$\sum_{k=1}^{r}\frac{1}{\sigma_k^2}  \geq \frac{r^2}{\sum_{k=1}^r
    \sigma_k^2}$ with equality if all singular values are equal, see e.g.\ \cite{xia1999proof}. A
  selection of basis functions that results in equal singular values
  can hence be regarded as \emph{variance optimal}.

\subsection{Bias}
The size of the bias contribution $\bm{\phi}(x')  \bm
   \theta_* - f_0(x')$
   in  \eqref{eq:prederror} depends on several factors. If we start by 
   assuming that the model is correctly specified, i.e.\  there exists a
   vector $\bm{\theta}_0$ such that for all $x$ we have 
     $f_0(x) = \bm{\phi}(x)  \bm
   \theta_0$ then the bias part of the error is 
     \begin{equation}
       \label{eq:21}
       \begin{aligned}
         \E \, e(x') & = \bm{\phi}(x') \bm \theta_* - f_0(x') =
         \bm{\phi}(x') \bm\Phi^+ \bm\Phi \bm\theta_0 - f_0(x')   \\ & =
         \bm{\phi}(x') ( \bm\Phi^+ \bm\Phi - \bm{I}) \bm\theta_0
       \end{aligned}
     \end{equation}
   \begin{itemize}
   \item
In the underparametrized case $(n\leq N)$, we see that $ \bm\Phi^+
     \bm\Phi - \bm{I} = 0$, since $\bm\Phi$ has full rank, thus the bias is zero.
 \item If $n>N$ then $\bm\Phi^+
     \bm\Phi - \bm{I} $ is a rank $n-N$ projection matrix which projects onto the
     nullspace  of $\bm\Phi$. The bias of $e(x')$ hence depends on the
     co-linearity of  the projected true parameter vector and the row
     vector(s) $\bm{\phi}(x')$. It also follows that the bias is zero
     if $\bm{\theta}_0 = \bm{\Phi}^T \bm{p}$ for some vector $\bm
     p$. In effect, this tells us that of all possible correctly
     specified models $f_0(x) = \bm{\phi}(x) \bm\theta_0$ only the
     $N$ dimensional subset of models given by  a parameter that can
     be expressed as $
     \bm{\Phi}^T \bm{p}$ will have zero bias since $(\bm\Phi^+
     \bm\Phi - \bm{I})\bm{\Phi}^T \bm{p}=0$. The set of zero bias
     models hence depends explicitly on the training data through the
     properties of the matrix $\bm\Phi$. As $n$ increases the
     size of the set of true functions with a non-zero bias also increases as the size
     of the nullspace of $\bm\Phi$ increases with $n$. 
   \end{itemize}
   For the misspecified case, when true function $f_0(x)$ is not part of the para\-metrized model class $f(x;\hat{\bm\theta)}=\bm\phi(x)\bm\theta$ the bias is
   \begin{equation}
     \label{eq:22}
     \E \, e(x') =   \bm{\phi}(x')  \bm \theta_* - f_0(x') =
     \bm{\phi}(x')  \bm\Phi^+ \bm{f}_0(\bm{x}) -f_0(x') 
   \end{equation}
  For the overparametrized case when $n\geq N$ the estimated model
  interpolates the training data. Hence, if $x'$ is very close to one
  of the training data inputs $x(t)$ in $\mathcal{D}$, (see \eqref{eq:Dset}), we can expect the bias
  $\E\,e(x')$ to be very small if the basis functions $\phi_i(x)$ are
  continuous.  In general when $x'$ is further away from the training
  samples the size of the bias error is difficult to characterize beyond
  the expression given in \eqref{eq:22}.     

\subsection{Ridge-Regression}
\label{sec:ridge-regression}
If we add a squared penalty of the parameters to the LS norm in
\eqref{eq:LSmatrix} we obtain
the ridge regression solution. For a positive scalar $\lambda$, the parameter estimate is
given by
  \begin{equation}
    \label{eq:ridge_regressin}
    \begin{aligned}
      \hat{ \bm \theta} & = \arg \min_{\bm\theta} \left\| \bm{y} -
        \bm{\Phi} \bm\theta \right\|^2 + \lambda \|\bm \theta \|^2\\ & =
      \arg \min_{\bm\theta} \left\| \ma{\bm{y} \\ \bm{0}} -
        \ma{\mathbf{\Phi} \\ \sqrt{\lambda} \mathbf{I}} \bm\theta
      \right\|^2
    \end{aligned}
\end{equation}
   When $\lambda>0$ the extended
   regression matrix always has full rank and hence the LS problem has a
   unique solution. Using the SVD of $\bm \Phi$ defined in
   \eqref{eq:PhiSvd} we can explicitly write
   the solution as
   \begin{equation}
     \label{eq:11}
     \begin{aligned}
       \hat{\bm \theta} & = \left( \ma{\mathbf{\Phi} \\ \sqrt{\lambda}
           \mathbf{I}} ^H \ma{\mathbf{\Phi} \\ \sqrt{\lambda}
           \mathbf{I}}\right)^{-1} \mathbf{\Phi}^H \bm y = \left(
         \mathbf{\Phi}^H\mathbf{\Phi} + \lambda \mathbf{I}
       \right)^{-1}\mathbf{\Phi}^H \bm y \\
       & = \left( \sum_{k=1}^{r} (\sigma_k^2+\lambda) \bm{v}_k
         \bm{v}_k^H + \sum_{k=r+1}^{n} \lambda \bm{v}_k \bm{v}_k^H
       \right)^{-1} \times \\ &  \left( \sum_{k=1}^{r} \sigma_k \bm{v}_k
         \bm{u}_k^H\right) \bm y
        = \sum_{k=1}^{r} \frac{\sigma_k}{\sigma_k^2+\lambda}
       \bm{v}_k \bm{u}_k^H \bm y 
     \end{aligned}
   \end{equation}
   where the sum $\sum_{k=r+1}^{n} \lambda \bm{v}_k
   \bm{v}_k^H$ vanishes if $n \leq N$.
   It is clear from \eqref{eq:11} that the ridge
   regression solution converges to the minimum norm solution \eqref{eq:min_norm_sol} as
   $\lambda\rightarrow 0$.

   Following the same analysis as above we have $\hat{\bm \theta} =
   \bm \theta_* + \tilde{\bm \theta}$ where the noise free solution is given by
   \begin{equation}
     \label{eq:13}
     \bm \theta_* = \sum_{k=1}^r \frac{\sigma_k}{\sigma^2_k + \lambda}\bm{v}_k
  \bm{u}_k^H \bm{f}_0(\bm{x})
   \end{equation}
   and the noise-induced error is given by
   \begin{equation}
     \label{eq:14}
     \tilde {\bm \theta} = \sum_{k=1}^r \frac{\sigma_k}{\sigma^2_k+ \lambda}\bm{v}_k
  \bm{u}_k^H \bm{z} 
\end{equation}
The expression on the variance of the model for a new value $x'$
corresponding to \eqref{eq:simp_pecov} is given by
\begin{equation}
  \label{eq:15}
     R_e(x') = r_z \sum_{k=1}^{r} \frac{\sigma_k^2}{(\sigma^2_k+\lambda)^2} (\bm{\phi}(x')
   \bm{v}_k) (\bm{\phi}(x') \bm{v}_k)^H 
\end{equation}
It is clear that the variance can be reduced by increasing $\lambda$.
However, if
$\lambda>0$ and $n\geq N$ then
\begin{equation}
  \label{eq:16}
  \begin{aligned}
    \bm \Phi \hat{\bm \theta} & = \left[\sum_{k=1}^r \sigma_k \bm{u}_k
      \bm{v}^H_k\right] \sum_{k=1}^{r}
    \frac{\sigma_k}{\sigma_k^2+\lambda} \bm{v}_k \bm{u}_k^H \bm y \\ &=
    \sum_{k=1}^{r} \frac{\sigma^2_k}{\sigma_k^2+\lambda} \bm{u}_k
    \bm{u}_k^H \bm y \not = \bm y
  \end{aligned}
     \end{equation}
which means that the estimated model does not  interpolate the
training data. This effect is commonly known as shrinkage since the
estimated model parameters are smaller in magnitude than the LS minimum norm
solution. The shrinkage effect will hence add to the total bias of the
estimated model.    

\subsection{Analysis of the smallest singular value}
\label{sec:analysis-if-smallest}
In this section we derive results on the behaviour of the smallest
singular value as the model order increases. The results gives a direct
explanation to the double descent phenomenon. We will show
\begin{itemize}
\item that when $n>N$ then the minimum singular
  value of $\bm \Phi$ for increasing model orders is non-decreasing.
\item that for $n<N$ then the minimum singular value
  is non-increasing for increasing model orders.
\end{itemize}
The result is based on the following general matrix result. 
\begin{thm}
  Let $\bm \Phi$ denote a matrix with $n$ columns and $N$ rows and define  $\bar{\bm \Phi} = \ma{\bm \Phi &
  \bm \phi_{n+1}}$ where $\bm \phi_{n+1}$ is an arbitrary vector.  
  \begin{enumerate}
  \item 
  Assume $n<N$ and  let $\sigma_1\geq \sigma_2 \geq \ldots \sigma_n$ denote the
  singular value of $\bm \Phi$ and let  $\bar \sigma_1\geq \bar \sigma_2 \geq
\ldots \bar \sigma_{n+1}$ denote the singular values of $\bar{\bm \Phi}$ . Then
\begin{equation}
  \label{eq:17}
  \bar \sigma_1 \geq \sigma_1 \geq \bar \sigma_2 \geq \sigma_2 \geq
  \ldots \geq \sigma_n \geq \bar \sigma_{n+1}
\end{equation}
\item Assume and $n \geq N$ and let $\sigma_1\geq \sigma_2 \geq \ldots \sigma_{N}$
denote the singular values of $\bm \Phi$ and let  $\bar \sigma_1\geq \bar \sigma_2 \geq
\ldots \bar \sigma_{N}$ denote the singular values of $\bar{\bm
  \Phi}$. Then
\begin{equation}
  \label{eq:18}
    \bar \sigma_1 \geq \sigma_1 \geq \bar \sigma_2 \geq \sigma_2 \geq
  \ldots \geq \bar \sigma_{N} \geq \sigma_{N}
\end{equation}
\end{enumerate}
\end{thm}
\begin{pf}\
  The result follows from \cite[Theorem 4.3.15]{Horn&Johnson:85} or
  \cite[Corollary 3.1.3]{Horn&Johnson:91}
\end{pf}
\begin{cor}\ 
  \begin{enumerate}
  \item If the model order satisfies $n<N$ then a model order increase
    will 
    result in a non-increase (unchanged or decreased size) of the smallest singular value.
  \item If the model order satisfies $n\geq N$ then a model order increase will
    result in a non-decrease (unchanged or increased size) of the smallest singular value.
  \end{enumerate}
\end{cor}
The presented result shows that if the inverse of the smallest singular
value has a maximum, then it will appear when $n=N$. As the level of
the variance is highly dependent on the smallest singular value as
shown in Section~\ref{sec:analysis} the maximum variance will in
general appear for model order equal to $N$ and the double descent
curve will peak at $n=N$ if the error is dominated by the variance
contribution. 
\subsection{Does overparametrization give any advantages?}
\label{sec:does-overp-give}
The key finding in the section above is that for $n>N$ the smallest
singular value $\sigma_r$ will not decrease with $n$. It will stay the
same or increase. For the
miss-specified case where the noise-free solution is given by
\begin{equation}
  \label{eq:23}
  \bm \theta_* = \bm{\phi}^+\bm{f}_0(\bm{x}) = \sum_{k=1}^r \frac{1}{\sigma_k}\bm{v}_k
  \bm{u}_k^H \bm{f}_0(\bm{x})
\end{equation}
we can conclude that the magnitude of the elements in $\bm \theta_*$
is highly influenced by the value of $1/\sigma_r$. As the model order
increases $1/\sigma_r$ will in general decrease and the magnitude of the elements in
$\bm \theta_*$ will decrease. From the predictive point of view of the
estimated models, i.e.\ values outside the training set, it seems more
natural that models with a smaller norm of the parameter vector are
better than models that have very large norm of the parameter vector. A second, more clear,
benefit is that the noise sensitivity is decreased as the model
order increases, at least for moderate values above $N$. This effect
is of course most pronounced when the regression matrix is close to
singular when $n=N$ and hence,  the
smallest singular value is closest to zero.

\section{Numerical illustrations}
\label{sec:numer-illustr}

In this section we examine some simulated numerical examples and
interprete the results with help of the results derived in the
analysis section. In all examples we will estimate regression models
with varying orders using the scalar valued (i.e.\ $p=1$) complex exponential $\phi(x)= e^{j2\pi f x}$ as
basis functions. A model structure is defined by the set of
frequencies $\mathcal{F} = \{f_i\}_{i=1}^n$ and is given by
\begin{equation}
  \label{eq:20}
   f(x;\bm\theta) = \sum_{i=1}^n \theta_i e^{j2\pi f_i x}.
 \end{equation}
 We generate the training data according to \eqref{eq:datagen} were we
 select $x(t)=t$ and let $t=0,1,\ldots,N-1$ where $N=10$ and the noise
 $z(t)$ are i.i.d.\ samples drawn from a zero mean circular symmetric complex Gaussian
 distribution with variance 0.1.  To evaluate the quality of an
 estimated regression model $f(x;\hat{\bm \theta})$ we derive the
 normalized mean square error of the predictions at the test
 data samples $x'(t)$ for $t=0,1,\ldots,N-1$
\begin{equation}
  \label{eq:7}
  \text{NMSE} = \frac{\sum_t | f_0(x'(t)) - f(x'(t);\hat {\bm \theta})|^2}
  {\sum_t | f_0(x'(t)) |^2}.
\end{equation}
We let  $x'(t) = t+\epsilon$, $t=0,\ldots,N-1$ and vary 
$\varepsilon$ in the different experiments. If $\varepsilon=0$ 
the estimated model is evaluated in the same data points as used for the learning.
If $\varepsilon=0.5$ we evaluate the quality of the model's
ability to predict values in between training samples, i.e.\ ability
to generalize.    

In the experiments we use two different model structures.
We set $n_\text{max} = 3N$ as the maximum model order.
For the model structure denoted as \emph{linear ordering} we define the
frequency set as
\begin{equation}
  \label{eq:24}
  \mathcal{F}_{\text{lin},n} = \{ k/n_\text{max} \}_{k=0}^{n-1}.
\end{equation}
For the model structure denoted as \emph{optimal ordering} we define the
frequency set as
\begin{equation}
  \label{eq:25}
  \mathcal{F}_{\text{opt},n} =
  \begin{cases}
    \{ k/N\}_{k=0}^{n-1}, & n\leq N \\
    \{ k/N\}_{k=0}^{N-1} \cup \mathcal{S}_{n-N}, & n>N
  \end{cases}
\end{equation}
where the set $\mathcal{S}_k$ is the first $k$ elements in the ordered set
$\{k/n_\text{max}\}_{k=0}^{n_\text{max}-1} - \{ k/N\}_{k=0}^{N-1} $.
If $n\leq N$ then the columns in the
regression matrix $\bm\Phi$ defined in \eqref{eq:Phi} are orthogonal to each
other and have equal norms. This in turn shows that all singular values are
non-zero and equal and hence this model structure is \emph{variance
  optimal} as discussed in Section~\ref{sec:analysis}.  We note that that 
$\mathcal{F}_{\text{lin},n_\text{max}} = \mathcal{F}_{\text{opt},n_\text{max}}$, i.e.\ the
two model structures are identical for $n=n_\text{max}$ but with a different ordering of the
basis functions. 

We will use the same type of basis functions to define two data generating
functions where the first one is given by  
\begin{equation}
  \label{eq:26}
  f_{0,\text{lin}} (x)  =    \sum_{k=1}^{10} \alpha_k e^{j 2 \pi \frac{k-1}{n_\text{max}} x}
\end{equation}
and the second one is 
\begin{equation}
  \label{eq:27}
    f_{0,\text{opt}} (x)   = \sum_{k=1}^{10} \alpha_k e^{j 2 \pi \frac{k-1}{N} x}
\end{equation}
In the Monte-Carlo Simulations below we will generate the coefficients $\alpha_k$
by sampling from a zero mean circular symmetric complex Gaussian distribution with unit variance.
The construction of the data generating systems implies that for
$f_{0,\text{lin}} (x)$ then all modell structures defined by
$\mathcal{F}_{\text{lin},n}$ for $n\geq 10$ will include the data
generating system in the model class. Along the same lines as above we notice
that  $f_{0,\text{opt}} (x)$ is included in all model structures
defined by the set $\mathcal{F}_{\text{opt},n}$ when $n\geq N$.
\begin{table}
  \centering
  \begin{tabular}{l|llll}
    Case & $\varepsilon$ & $f_0$ &  $n_\text{max}$ & N \\ \hline 
      A   & 0.5 & $f_{0,\text{lin}} (x)$ & 30 & 10 \\
      B   & 0   & $f_{0,\text{lin}} (x)$ & 30 & 10 \\
      C   & 0.5 & $f_{0,\text{opt}} (x)$ &  30 & 10 \\
      D   & 0 & $f_{0,\text{opt}} (x)$ & 30 & 10 
  \end{tabular}
  \caption{Definition of the different cases in the numerical examples}
  \label{tab:exp}
\end{table}
In Table~\ref{tab:exp} we define four experimental cases. For
each case we generate a data generating system using the
function according the $f_0$ column and create 500 training
datasets with added noise and 500 data sets without noise. For each dataset a model is estimated and the NMSE is
evaluated at the $x$ values defined by $t+\varepsilon$ for $t=0,\ldots,N-1$. 

\begin{figure}
  \centering
  \includegraphics[width=1.0\columnwidth]{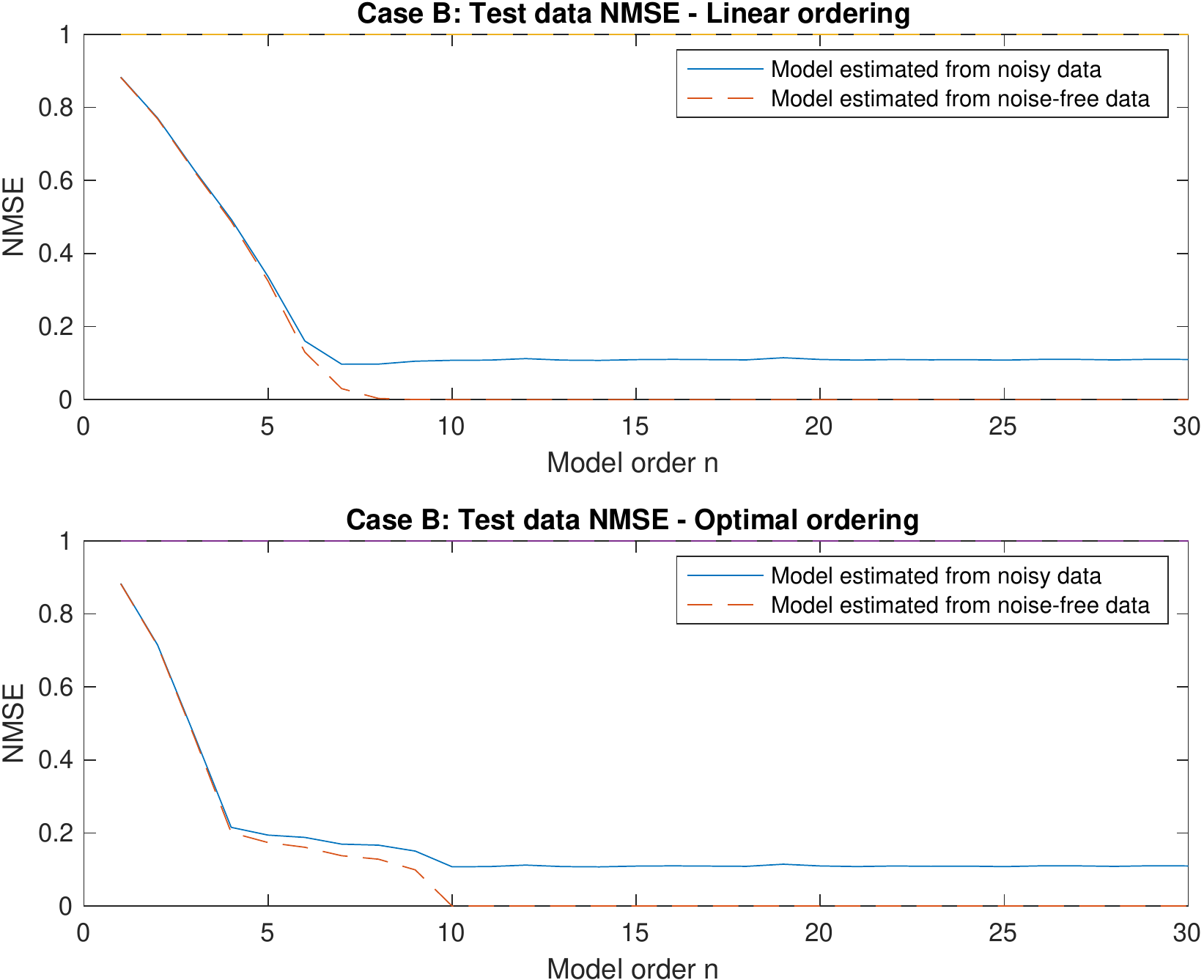}
  \caption{Case B: Graphs show normalized mean squared error as a
    function of model order for the linear ordering model structure
    (top) and the optimal ordering (bottom). }
\label{fig:B}
\end{figure}

\begin{figure}
  \centering
  \includegraphics[width=1.0\columnwidth]{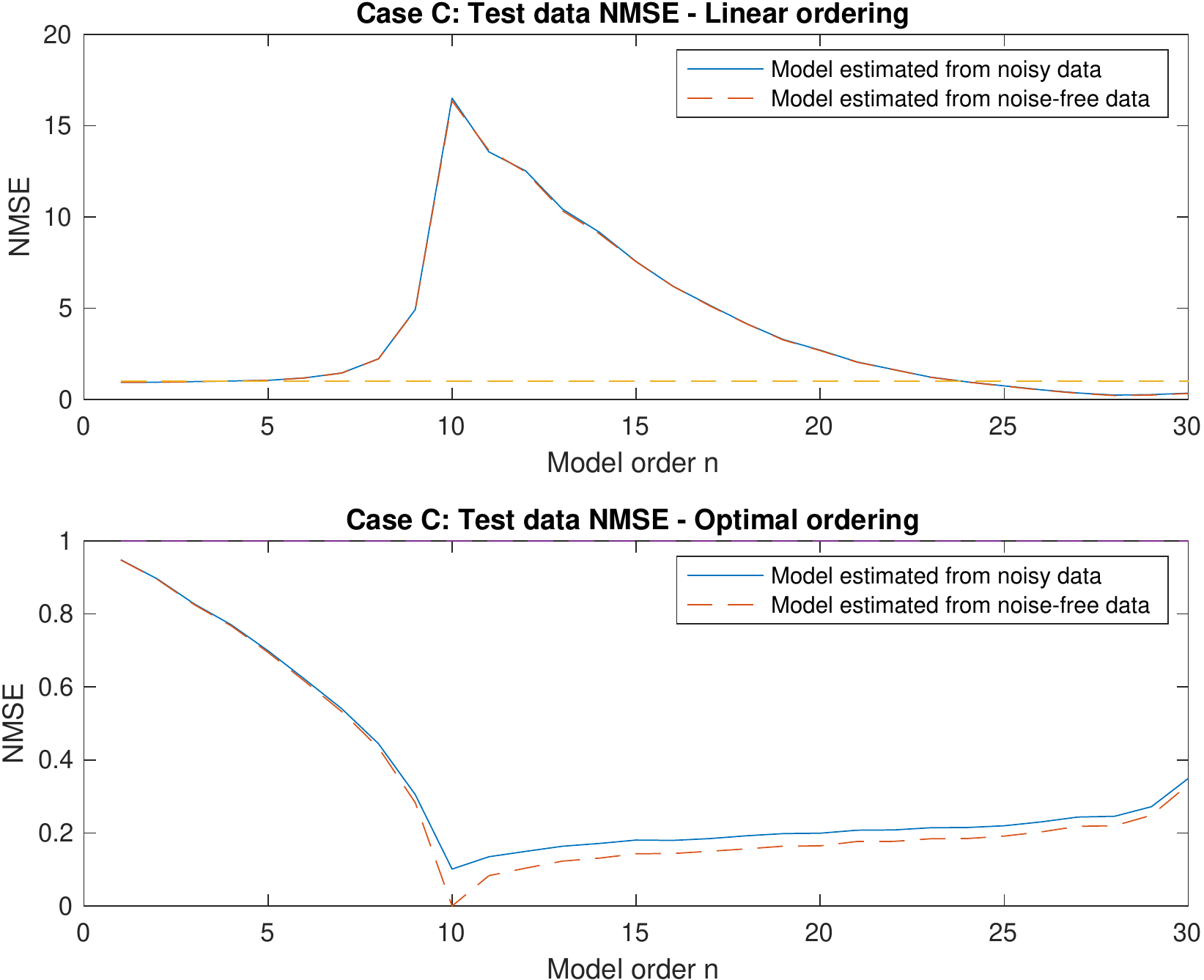}
  \caption{Case C: Graphs show normalized mean squared error as a
    function of model order for the linear ordering model structure
    (top) and the optimal ordering (bottom). }
\label{fig:C}
\end{figure}

\begin{figure}
  \centering
  \includegraphics[width=1.0\columnwidth]{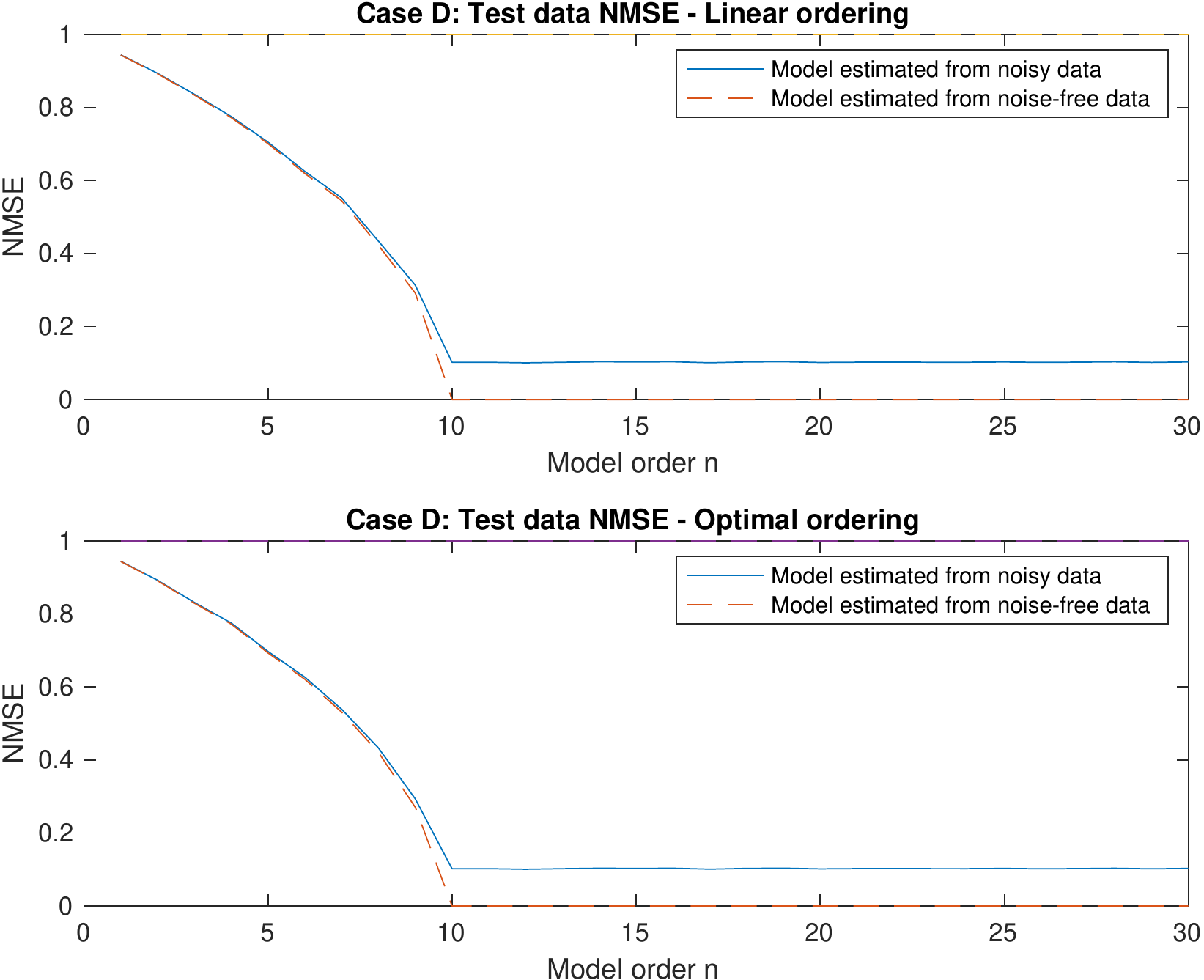}
  \caption{Case D: Graphs show normalized mean squared error as a
    function of model order for the linear ordering model structure
    (top) and the optimal ordering (bottom). }
\label{fig:D}
\end{figure}

The average NMSE over the Monte-Carlo simulations for the two model
structures as a function of the
model orders are reported in the graphs in 
Figure~\ref{fig:A} to Figure~\ref{fig:D}. In Figure~\ref{fig:SVD} the
inverse of the smallest singular value of
the regression matrix $\bm\Phi$ is illustrated as a function of model
order for the two model structures. 

\subsection{Discussion}
\label{Discussion}
The NMSE testing error is in the figures shown for models trained on
noise free data as well as trained on the noisy data. The testing
results on models trained on noise free data give direct information
on the NMSE caused by the bias contribution given by
\eqref{eq:21} and \eqref{eq:22}.  The testing
results on models trained on noisy data give information
about the total MSE caused by the bias contribution and
the variance contribution given by
\eqref{eq:simp_pecov}. 

\begin{figure}
  \centering
  \includegraphics[width=1.0\columnwidth]{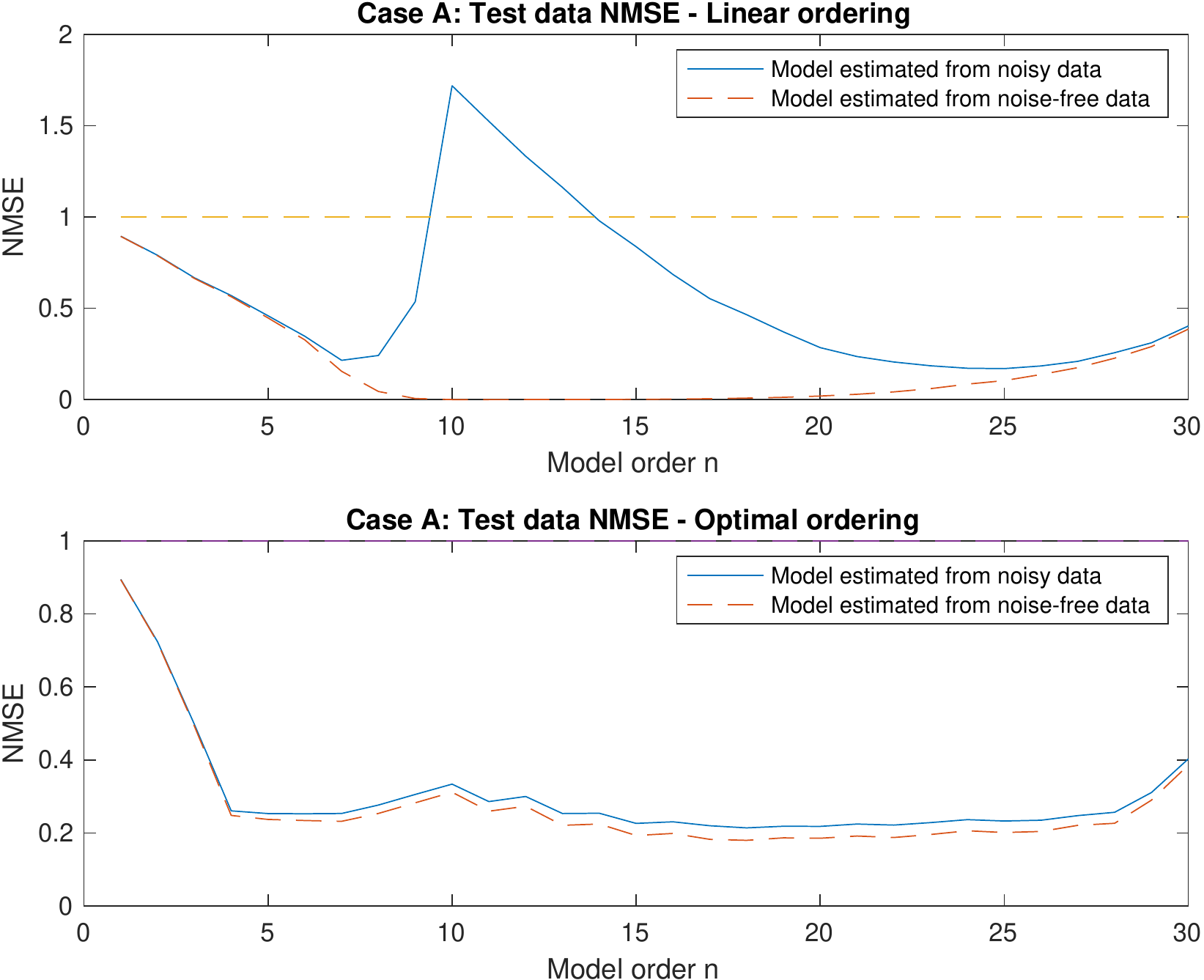}
  \caption{Case A: Graphs show normalized mean squared error as a
    function of model order for the linear ordering model structure
    (top) and the optimal ordering (bottom). }
\label{fig:A}
\end{figure}

\begin{figure}
  \centering
  \includegraphics[width=1.0\columnwidth]{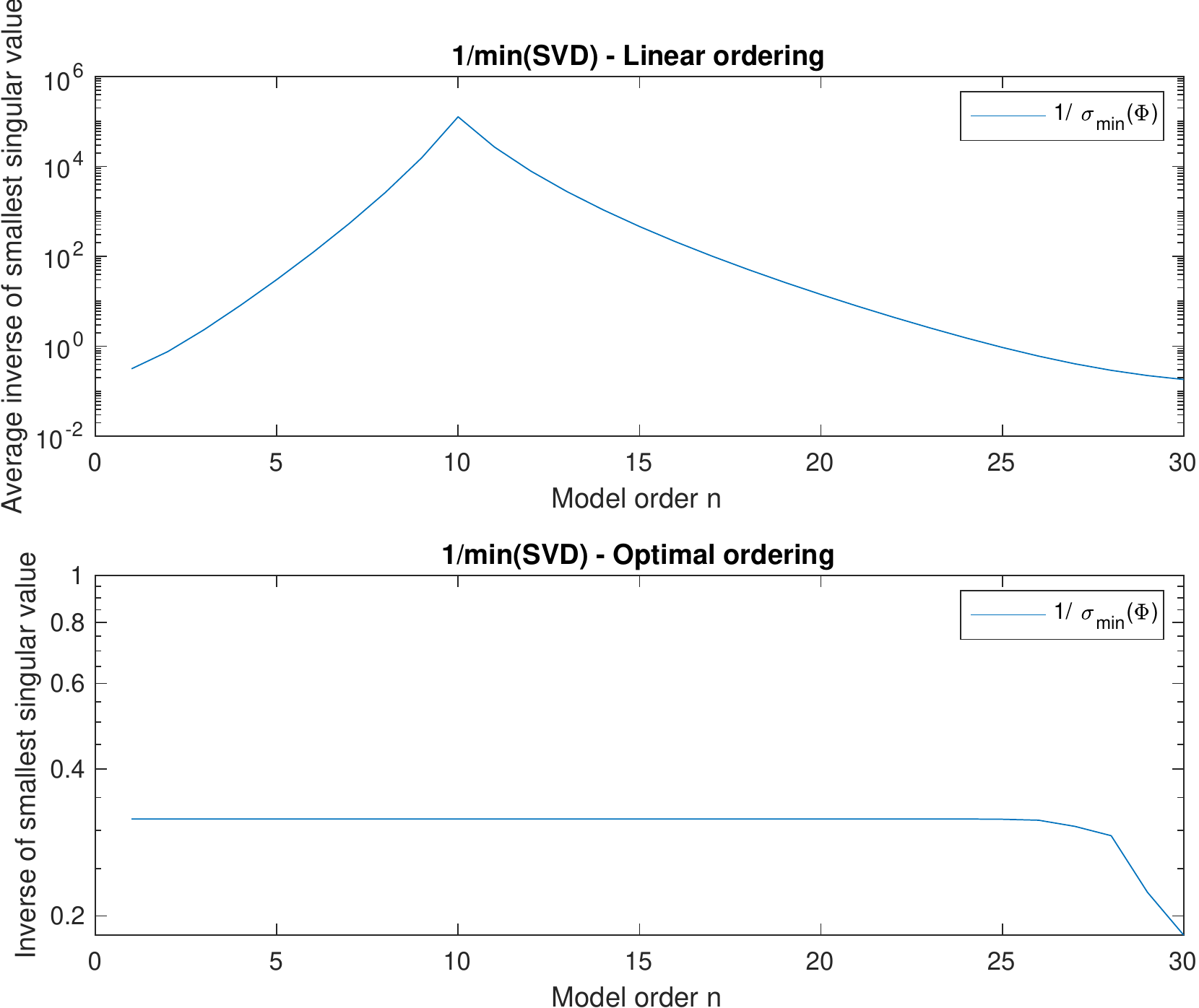}
  \caption{The graphs show the inverse of the smallest singular value
    for the regressor matrix $\bm \Phi$ as a
    function of model order for the linear ordering model structure
    (top) and the optimal ordering (bottom). }
\label{fig:SVD}
\end{figure}

In Case A the true system is in the linear ordering model class for
model orders $n\geq 10$. Hence for noise free data and $n=10$ we
recover the true model as seen in Figure~\ref{fig:A}. For the noise
free case the test data error has an increase again for model orders
larger than 20. This is the effect when the true model parameters are
not in the row space of $\bm\Phi$ as discussed below \eqref{eq:21}.
For models estimated from noisy data the double descent phenomenon is
clearly visible. A comparison with the top graph in
Figure~\ref{fig:SVD} show the qualitative agreement between the NMSE
and the inverse of the smallest singular value. The peaks are located
for $n=N=10$ in both graphs and the behaviour for the singular values
are in agreement with Corollary 2.  The result for Case A for the
optimal ordering model class is shown in the bottom graph in
Figure~\ref{fig:A}. For this case the true model is in the model set
for $n\geq 16$. Hence, even for noise free data this model structure
has a non-zero error that for the highest model orders increases again
for the same resons as discussed before. However, for the noisy case
the performance is significantly improved for this model structure and
is in par with the performance of the model estimated from the noise
free data. The reason for this is found in the bottom graph in
Figure~\ref{fig:SVD}. The inverse of the smallest singular value is
much smaller than the linear ordering model structure for all model
orders.  This implies that the variance as given by
\eqref{eq:simp_pecov} is much smaller as compared with the other model
structure. In Case B the same setup is used except that the test data
points are the same as the training data points. For the noise free
case we obtain zero error for both model structures for $n\geq
10$. For the noisy data case an error which is equal to the noise
level is obtained since all models interpolate the training data when $n\geq
10$. For case C an D the true system is now given by the optimal order
data structure. Hence, for $n=10$ the optimal ordering model structure
recovers the true model for noise free data and give the best NMSE for
the noisy data. For higher model orders the performance is slightly
reduced which again is attributed to an increase in the bias as
discussed before. For the linear ordering model structure it is only
at $n=28$ the true system is in the model class and it is at this
model order the best NMSE on test data are achieved. The
ill-conditioning of this model structure is for the lower model orders
clearly visible and the NMSE again has a peak at $n=N=10$.  For this
model structure we can conclude that overparametrization produces a
model with resonable performance as compared to the solutions for
model orders around 10.


\section{Conclusions}
\label{sec:conclusions}

The existence of  a double descent behaviour is closely related
  to the inverse of the smallest singular value of the  associated
  regression matrix. A model structure with a near singular regression matrix when $n=N$
  results in a double descent behavior for the NMSE on test data at
  other locations than the training data.

  To estimate overparametrized models, i.e. more parameters than
  training data using the pseudo inverse solution can be resonable
  (NMSR$<1$)  if the true parameter is close to the row space of the regression
  matrix. If this is not the case the solutions will have poor
  performance.

  To obtain robust overparametrized  solutions it is important to select a model
  class such that the minimum singular value of the associated
  regression matrix is as large as possible.  

\section*{Acknowledgement}
   The author would like to thank Daniel McKelvey for giving valuable comments
   on the manuscript and the reviewers for their constructive comments.

\bibliographystyle{IEEEtranS}

\end{document}